# Omega Model for Human Detection and Counting for application in Smart Surveillance System

Subra mukherjee
Assam Don Bosco University
Guwahati, India

Karen Das
Assam Don Bosco University
Guwahati, India

*Abstract*— Driven by the significant advancements in technology and social issues such as security management, there is a strong need for Smart Surveillance System in our society today. One of the key features of a Smart Surveillance System is efficient human detection and counting such that the system can decide and label events on its own. In this paper we propose a new, novel and robust model: *"The Omega Model"*, for detecting and counting human beings present in the scene. The proposed model employs a set of four distinct descriptors for identifying the unique features of the head, neck and shoulder regions of a person. This unique head-neck-shoulder signature given by the Omega Model exploits the challenges such as inter person variations in size and shape of people's head, neck and shoulder regions to achieve robust detection of human beings even under partial occlusion, dynamically changing background and varying illumination conditions. After experimentation we observe and analyze the influences of each of the four descriptors on the system performance and computation speed and conclude that a weight based decision making system produces the best results. Evaluation results on a number of images indicate the validation of our method in actual situation.

*Keywords - Omega Model; Human Detection; Surveillance; Back ground Subtraction; Gaussian Mixture Model (GMM); Mixture of Gaussians (MOG).*

## I. INTRODUCTION

The state-of-art of surveillance has made a quantum jump in recent years. However with the increase amount of video data to be processed it is becoming more and more unmanageable for human beings to monitor continuously. So if we could develop a surveillance system which could detect and classify objects, take decisions and label events autonomously, then a complete revolution can be brought in the current surveillance system. Vision based Human detection and counting is currently one of the most challenging tasks in the field of computer vision. The general surveillance cameras are like machines that can only see, but cannot decide or identify things or events on its own. So, keeping in mind the present day scenarios, it is important that we make our surveillance system intelligent and smart. Therefore, we propose to design a new framework to robustly and efficiently detect and count human beings, for application in surveillance. The proposed system would consist of: Background subtraction, boundary extraction, Head-neck-shoulder detection and, finally human/non-human classification and based on that, counting the number of human present in a scene. For these, we first intend to subtract the background and extract the foreground of any real time video. There are a lot of techniques available for background subtraction. And Gaussian Mixture Model (GMM) is found to be more efficient in the literature. So we intend to use GMM for this purpose. Moreover some of the commonly faced problems in background subtraction are sudden changes in illumination, dynamic background, camouflage, etc. Hence we intend to design a robust adaptive GMM algorithm which can effectively deal with all these problems and produce a foreground mask. Secondly we intend to detect human presence in the scene by detecting the head and shoulder portion by using the *Omega Model*. We propose this model because the head-shoulder portion is the most unvarying part of human body. Based on the number of human beings detected we shall count the total number of human present in the scene. And hence the entire system could be used for application in an effective surveillance system.

The rest of the paper is organized as follows: In section II we discuss some of the related work in this field; in section III we give an overview of the method adopted for our work. In section IV our human detection and counting system is discussed giving a detailed description of our proposed Omega model explaining each of the descriptors and the algorithm. In section V we have explained results followed by conclusion and our future work in section VI.

## II. RELATED WORK

There is an extensive literature on shape classification. Various approaches for shape based classification are discussed in [1-10]. However different moving objects like bird, vehicle, etc may be present in the scene, so it is very important that we correctly distinguish humans from other moving objects. There are mainly two methods for classifying a moving object: shaped based detection and motion based detection [11]. In former one, human can be detected with the help of their shape information. This kind of a work was done in [12-14] where they used an SVM classifier to detect human beings based on finding people's head by searching for circular patterns through a 2D correlation using a bank of annular patterns. Also it is a general fact that non articulated human motion exhibits certain periodicity. This property was used by many researchers to detect human beings based on their motion. In [14] based on the color object's moving and background subtraction method, a color classifier based on the HS thresholds was proposed to detect moving object. In [15] edge-based features combined along with color and texture information was used for efficient human detection. In [16] human had been detected by detecting skin like pixels and locating each face like region. Also some researchers have employed model based human detection [17,18].In [17] such kind of work was done wherein they proposed a method for human detection by modeling human as flexible assemblies of parts represented by co-occurrence of local features. In [18] part detectors were learned by boosting a number of weak classifiers based on edgelet features. Recently in 2013 authors [19] have presented a method for human detection in range images captured from a vertically oriented camera by analysis of 3D range data.

### III. OVERVIEW OF THE METHOD

This section gives an overview of the method adopted in our Human detection and Counting system. One of the major challenges in the field of object recognition is the ability to detect human beings irrespective of the variations in pose, body shape, clothing, illumination, moving cameras and changing background. So in this work we have developed the Omega model that could detect human beings under all this challenging scenarios. The methodology or the general flow diagram of our work is shown below:

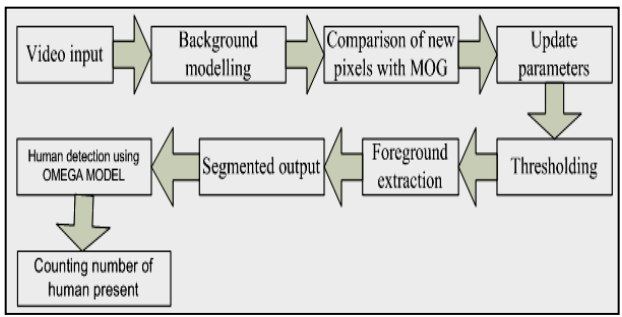

Fig. 1. General Flow Diagram Of Our Human Detection And Counting System.

The various steps involved are:

- Acquiring real time video input from any video acquisation device
- Background modelling using adaptive GMM
- Background subtraction and shadow detection
- Human detection and counting using the Omega Model.

### IV. HUMAN DETECTION AND COUNTING SYSTEM

In this work, we first perform adaptive background modeling to extract the foreground region from a real time Surveillance video. Then we acquire a set of these foreground images from a surveillance video. As we know that the human head shoulder portion is the most unvarying part of human body, so we have used this dominant feature as the key information and developed the Omega Model for human detection.

#### A. Foreground Extraction

A good surveillance system requires an accurate segmentation of moving objects from a video sequence. Foreground extraction is generally done by using background subtraction, optical flow and frame differencing. However,

Background subtraction is one of the most efficient and widely used methods for segmenting dynamic scene in a video. The most common paradigm for background subtraction is to use an explicit model of the background. Background is generally modeled based on some regular statistical characteristics. Intruding objects are then detected by comparing the statistical parameters of the modeled background with that of the current frame. However this method does not work well in surveillance scenarios where the background is generally subjected to challenges like dynamic lightning conditions, long term scene changes, bimodal background, repetitive flickering motions etc. So, for application in surveillance it is important that the parameters of the background are also adaptive. Hence we have employed the adaptive GMM method proposed in [20] for modeling the background.

*a) Gaussian mixture Model:*

A Gaussian Mixture Model (GMM) is a parametric probability density function represented as a weighted sum of Gaussian component densities. GMMs are commonly used as a parametric model of the probability distribution of continuous measurements or features in a biometric system, such as vocal-tract related spectral features in a speaker recognition system. GMM parameters are estimated from training data using the iterative Expectation Maximization (EM) algorithm or Maximum A *Posteriori* (MAP) estimation from a well-trained prior model.

A Gaussian mixture model is a weighted sum of M component Gaussian densities as given by the equation,

$$p(X|\lambda) = \sum_{i=1}^{M} w_i g(X|\mu_i, \Sigma_i)$$

where x is a D-dimensional continuous-valued data vector (i.e. measurement or features), $w_i$, i=1,...,M, are the mixture weights, and $g(x|\mu_i,\Sigma_i)$, i = 1,...,M, are the component Gaussian densities. Each component density is a D-variate Gaussian function of the form,

$$g(X|\mu_i, \Sigma_i) = \frac{1}{(2\pi)^{\frac{D}{2}} |\Sigma_i|^{1/2}} \exp\left\{-\frac{1}{2}(X - \mu_i)' \Sigma_i^{-1}(X - \mu_i)\right\}$$

with mean vector $\mu_i$ and covariance matrix $\Sigma_i$. The mixture weights satisfy the constraint that $\sum_{i=1}^{M} w_i = 1$.

*b) Parameter updates*

The new pixel value $Z_t$ is checked against each Gaussian. A Gaussian is labeled as matched if

$$\|Z - \mu_h\| < d\sigma_h$$

Then its parameters may be updated as follows:

$$w_{i,t} = (1 - \alpha) * w_{i,t-1} + \alpha * M_{i,t}$$
$$\mu_t = (1 - \rho) * \mu_{t-1} + \rho * Z_t$$
$$\sigma_t^2 = (1 - \rho) * \sigma_{t-1}^2 + \rho * (Z_t - \mu_t)^T * (Z_t - \mu_t)$$
$$\rho = \alpha * N(\mu_t)$$

Where α is the learning rate for the weights.

If a Gaussian is labeled as unmatched only its weight is decreased as

$$w_{i,t} = (1 - \alpha) * w_{i,t-1}$$

If none of the Gaussians match, the one with the lowest weight is replaced with $Z_t$ as mean and a high initial standard deviation.

The rank of a Gaussian is defined as w/σ. This value gets higher if the distribution has low standard deviation and it has matched many times. When the Gaussians are sorted in a list by decreasing value of rank, the first is more likely to be background. The first B Gaussians that satisfy (1) are thought to represent the background.

$$B = \arg\min_b \left( \sum_{k=1}^{b} w_i > T \right) \quad (1)$$

The Gaussian mixture model (GMM) is adaptive; it can incorporate slow illumination changes and the removal and addition of objects into the background. Further it can handle repetitive background changes like swaying branches, a flickering computer monitor etc. The higher the value of T in (1), the higher is the probability of a multi-modal background.

In our work we have modeled the background as a mixture of three Gaussians.

*B. Omega Model for Human Detection*

Significant research has been devoted to detecting people in images and videos. Human detection is a challenging classification problem which has many potential applications in the field of machine vision. The main problems in detecting human beings are due to the variations in pose, body shape, clothing, illumination, moving cameras and changing background.

Therefore the main challenge is to find a set of unique features that characterizes human being in a scene, while remaining resistant to the above mentioned problems.

Thus in this work a new algorithm is presented to detect human beings in still images using a set of four descriptors. After the foreground extraction, the human beings have been detected by studying some of their invariant features like the head-neck- shoulder signature.

*a) Outline of approach for Human Detection system:*

The block diagram of the proposed Human detection system is as shown below in Fig2.

This approach uses a shape based representation of the extracted foreground contour for human detection. The advantages of this approach are:

- It can detect human beings even in partial occlusion (when legs are partially occluded).
- It is tolerant to varying human pose.
- It can detect human beings even if the person is not facing the camera directly.

- The final decision is weight based and depends on multiple evidences obtained from descriptors.

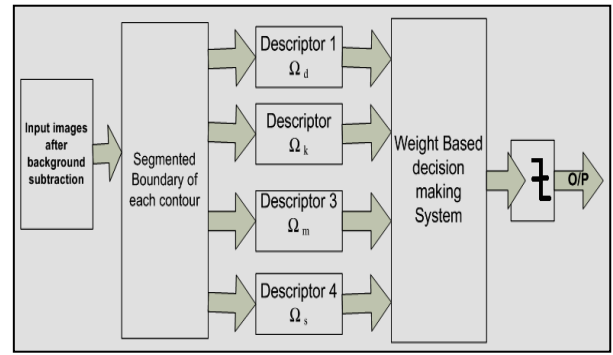

Fig. 2. Flow chart for Omega model for human detection

In this approach, the boundary of the contour of the extracted foreground object has been examined experimentally to obtain some of the invariant features of human beings from the shape of the contour. Four descriptors have been designed to specifically analyze these invariant features and thereafter take a weight based decision to detect the presence of human beings in the scene.

*b) Descriptors for Human Detection:*

The choice of the distinguishing features for classification is a critical design step and depends on the characteristics of the problem domain. Having extracted the contour of the foreground objects, a set of invariant features have been chosen to detect the presence of human being in the scene. In this work we have developed four descriptors to classify the human beings from other non- human objects by using distinct features that are simple to extract as well as invariant to irrelevant transformations.

From the set of boundary points obtained, by processing the contour of the segmented objects, the main aim here is to develop descriptors that describe the '*Omega*' shape (i.e. the shape of upper portion of human body) in the best possible manner.

The four Descriptors we use are as follows:

**Descriptor 1 ($\Omega_d$) :** ( Head-neck- shoulder dimensions of $\Omega$)

- This shape based descriptor is firstly defined by its dimension given as shown in figure (3(b)):

  {$Y_{max} - Y_{min}, X_{min} - X_{max}$}

- A bounding box is designed to include the object of interest and whose axes are aligned with the image axes as shown in figure (d)
- Based on the set of boundary points obtained, co-ordinates of the centroid are calculated.
- From this obtained centroid, data for width of shoulder and neck is obtained.
- The data obtained is then experimentally analyzed with a number of training images to obtain a threshold for

describing the optimum ratio of these width and compare with the testing images.

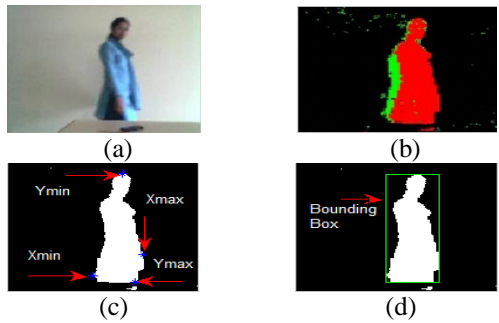

Fig. 3. (a) original image, (b) image after background subtraction (c) dimensions of the image (d) Bounding box for the contour.

Based on this threshold (obtained experimentally) a decision is made if a human being is present in the scene or not.

**Descriptor 2 ($\Omega_m$)**: *(Radial Feature of $\Omega$)*

- This descriptor particularly defines the radial feature of the human head.
- Based on experimental analysis the upper (head) portion of the contour is extracted and a point (S') lying somewhere between the neck and tip of head is obtained.
- The radial distance between each of the points in the boundary and point S' is calculated.
- The pattern of occurrence of these distances is observed for human contours.
- Based on the pattern a decision is taken if the extracted contour is that of human head or not.

**Descriptor 3 ($\Omega_k$)**: *(Curvature of $\Omega$)*

- This descriptor classifies human based on the information of curvature of human head-neck-shoulder portion.
- At each point in the boundary of the contour, curvature is estimated which is an indicator of the amount of bending of the curve that occurs at that position.
- Based on the set of curvature values obtained for each of the boundaries of the contour, the patterns have been studied.
- Analysis of these pattern shows that a specific number of local minimas occur if the contour under observation is that of human being.
- Based on this experimental analysis, a threshold is obtained and decision is taken whether a human being is present in the scene or not.

**Descriptor 4 ($\Omega_s$)**: *(Convexity of $\Omega$)*
- Here shape description is based upon the convex hulls of the set of boundary points obtained from the extracted contour.

- The convex hull of the set of boundary points of the contour is the enclosing convex polygon with the smallest possible area.
- So here we analyze the convexity of the head-shoulder portion of human body. We define convexity ($R_s$) as:

$$R_s = \frac{\text{area of rectangle bounding the upper segmented contour}}{\text{area of Convex hull}}$$

- The ratio obtained above have been analyzed for a number of test image and based on experimentation a threshold have been obtained to detect the presence of human being in the scene.

Weights are assigned to each of the descriptor based on the experimental analysis. Finally based on the decisions obtained from the four descriptors, a weight based decision is taken and if outcome is above a certain threshold than a human being is said to be present in the scene.

The complete algorithm for detecting human beings employing these four descriptors is given in the next sub-section.

*c) Algorithm for Human Detection:* The algorithm, that have been designed for human detection is as described below.

Each of the extracted contour present in an image is processed to find human beings based on the descriptors. Each of the descriptors has been assigned a weight depending on their performance analysis. Finally a weight based decision is taken and compared with a standard threshold ($\Omega_{th}$) obtained from experimental analysis and accordingly the human beings present in the scene is detected and counted.

V. EXPERIMENTAL RESULTS AND DISCUSIONS

The results obtained for human detection and counting are very satisfactory. Here after background subtraction, the contours present in the segmented foreground image is processed using the developed algorithm for *Omega model*. The resolution of the camera used in the work is 120X160, running in a 32 bit operating system, 2.00 GHz processor, and 2 GB RAM. The achieved speed of execution for foreground extraction is 21 fps. The developed algorithm was then tested on 100 frames, each consisting one or more number of human beings (including frames where the human is partially occluded i.e legs are occluded) and 50 frames that did not contain human beings. We have achieved a success rate of 95%. The time required to detect human in a frame is 18ms. Certain error arouse due to complete occlusion of the head shoulder portion. However our method is tolerant to changing background and also effectively deals with different poses of head- shoulder shape taken from different camera angles. A Matlab based tool with Graphical User Interface (GUI) has been developed for the ease of use by anyone to detect the number of human being present at a scene.

Algorithm for Omega model for human detection:

**Descriptor 1($\Omega_d$)** (*Neck shoulder dimensions of $\Omega$*)

1. Get the boundary points $\{x_i, y_i\}$ for each contour obtained from background subtraction.
2. Find $Y_{min}, Y_{max}, X_{min}, X_{max}$ values for each of the boundaries obtained in step1.( refer fig.3(c) )
3. Obtain the height (**h**) and width (**w**) of the contour.
4. Find the co-ordinates of centroid ($C_x, C_y$) of the contour.
5. Find distance, **d**= 1/3 of **h** and **d'** =1/2 of **d**.
6. Obtain the following points:
    (a)  $X_{min1}, X_{max1} < C_x$
    (b)  $X_{min2}, X_{max2} > C_x$
7. Define two variables $ш_1$ (neck width) and $ш_2$ (shoulder width) such that:
    (a) $ш_1 = X_{max1} - X_{min2}$
    (b) $ш_2 = X_{max2} - X_{min1}$
8. Take a decision, $[\Omega_d = 1$ if $ш_1/ ш_2 = T_d,$
    $=0$ otherwise].

**Descritor2 ($\Omega_m$)**: *(Radial feature of $\Omega$)*

9. Obtain a point '**S**' (from experimentation) lying between nose and neck in the y direction from $C_y$ to $Y_{min}$.
10. Find the distance **S'** between S and $Y_{min}$.
11. Define a set of points $\{S_1', S_2', S_3', S_4', S_5', S_6' \ldots\ldots\ldots S_n'\}$ to all other points in the segmented boundary.
12. Take a decision such that:
    $[\Omega_m =1$ if $S' > \{S_1', S_2', S_3', S_4', S_5', S_6' \ldots\ldots\ldots S_n'\},$
    $=0$ otherwise].

**Descriptor 3($\Omega_k$):** *(Curvature of $\Omega$)*
13. Calculate the absolute values of curvatures $\{C_i\}$ for the segmented boundary (1/3 of **h**) given as
$$C = \frac{x'y'' - x''y'}{\sqrt[3]{x'^2 - y'^2}}$$
14. Find the number of local minima for the segmented upper contour.
15. Take a decision:
    $[\Omega_k = 1,$ if $a_1 < C < a_2$
    $= 0$ otherwise], where $a_1$ and $a_2$ are the thresholds for number of local minima.

**Descriptor 4($\Omega_s$)** *(Convexity of $\Omega$)*

16. Find the convex hull of the upper segmented boundary of the contour.
17. Find the area ($A_c$) of the convex hull.
18. Find the area ($A_r$) of the rectangle bounding the upper segmented contour.
19. Find the ratio: $R_s = A_r/ A_c$.
20. Take a decision:
    $[\Omega_s = 1,$ if $r_1 < R_s < r_2$
    $= 0$ otherwise],
    where $r_1$ and $r_2$ are the thresholds values for $R_s$.
21. Finally take a weight based decision.

Define a function $H(\Omega)$ such that:

$H(\Omega) = S_d\Omega_d + S_m\Omega_m + S_k\Omega_k + S_s\Omega_s$,

where $S_d, S_m, S_k, S_s$ are the respective weights of descriptors.
An extracted contour is said to be that of human if

$H(\Omega) >= \Omega_{th}.$

When the tool is started, the user can browse and select any image containing contours of foreground object. The GUI shows the user a bounding box for each of the object present in the original image. Then the segmented contour for each of the object is also shown in the window and then using the algorithm it automatically generates the count of the number of human beings present in the image. A screen shot of the developed GUI is as shown below.

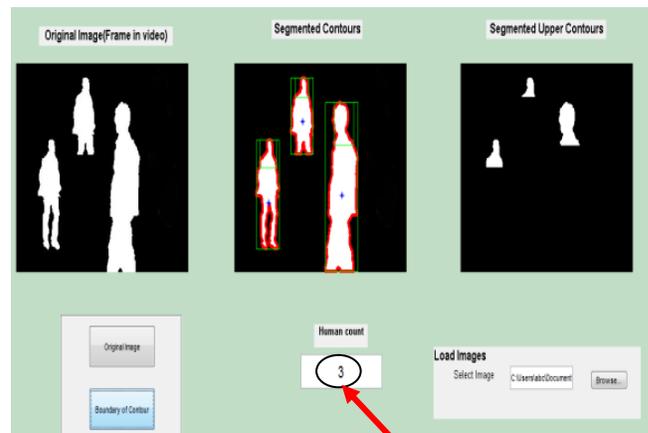

*(i) Human count=3*

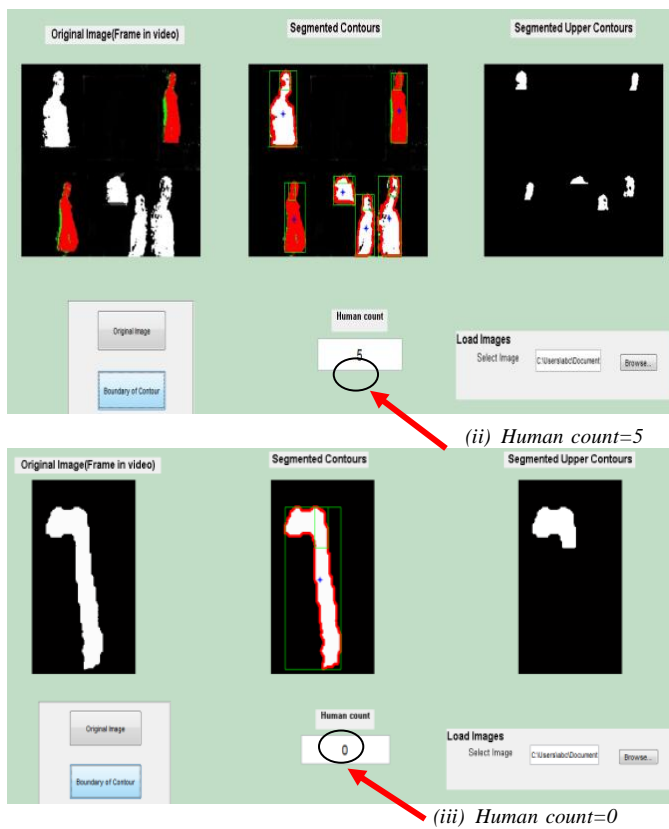

*(ii) Human count=5*

*(iii) Human count=0*

Fig. 4. GUI shown for human count in images using the developed algorithm.

## VI. CONCLUSION AND FUTURE WORK

A method for human detection and counting has been presented in this paper. The key feature of our work is, we have employed four descriptors to detect four invariant and significant feature of human head-shoulder region to achieve our goal. We studied the influence of various descriptor parameters and conclude that none of them can individually detect human, hence we employed a weight based decision system for a good performance. Experiments performed on several images validate the effectiveness of our approach.

In our future work we shall focus on implementing the Omega Model in video, and hence develop a Real-time Smart Surveillance Systems that can decide and label events and give threat alerts for security conscious venues.